\documentclass[prl, twocolumn, showpacs, superscriptaddress]{revtex4-1}
\usepackage{amsmath}
\usepackage{mathrsfs}
\usepackage{amsfonts}
\usepackage{amssymb}
\usepackage{graphicx}
\usepackage{subfigure}
\usepackage{amsthm}
\newtheorem{definition}{Definition}

\begin{document}

\title{Equivalence in Deep Neural Networks via Conjugate Matrix Ensembles}

\author{Mehmet S\"uzen}
\email[]{suzen@acm.org;mehmet.suzen@physics.org}


\begin{abstract}
A numerical approach is developed for detecting the equivalence of deep learning architectures. The method is based on generating Mixed Matrix Ensembles (MMEs) out of deep neural network weight matrices and {\it conjugate circular ensemble} matching the neural architecture topology. Following this, the empirical evidence supports the {\it phenomenon} that difference between spectral densities of neural architectures and corresponding {\it conjugate circular ensemble} are vanishing with different decay rates at the long positive tail part of the spectrum i.e., cumulative Circular Spectral Difference (CSD). This finding can be used in establishing equivalences among different neural architectures via analysis of fluctuations in CSD. We investigated this phenomenon for a wide range of deep learning vision architectures and with circular ensembles originating from statistical quantum mechanics. Practical implications of the proposed method for artificial and natural neural architectures discussed such as the possibility of using the approach in Neural Architecture Search (NAS) and classification of biological neural networks.
\end{abstract}


\maketitle

\section{Introduction}

Constructing equivalence relations among different mathematical structures 
are probably one of the most foundational concepts in sciences \cite{rosser2008a}, 
a practical interest as well, beyond being a theoretical building block of many
quantitative fields. This manifests in many fields of physical sciences and in practice, 
such as for equivalence of graph network ensembles \cite{barre2007a, den2018a}, single-molecule 
experiments \cite{costeniuc2005a, suzen2009a}, Bayesian Networks \cite{chickering2002}, 
between ranking algorithms \cite{ertekin2011a} and Brain network motifs \cite{sporns2004motifs, bullmore2009a}.

Recent success of deep learning \cite{schmidhuber15a, lecun2015a} in different learning
tasks showing skills exceeding human capacity, specially in vision tasks brings 
the need for both understanding of these systems and build neural architectures
in an efficient manner. In this direction, detecting equivalent deep learning architecures
are not only interesting for theoretical understanding but also for finding more
efficient architecture as a design principle. Neural Architecture Search (NAS) \cite{elsken19a, ren2020a}
or search for smaller equivelent network, i.e., architecture compression \cite{kumar19a, serra2020a} 
are valuable tools in achieving this aim.

Approach in establishing equivalence between two deep learning architectures are taken here lies
in the analysis of eigenvalue spectra of the trained weights, network topology and generating
{\it conjugate random matrix ensemble}. There will be no dependence on the activation functions or training 
procedure in establishing such equivalance. This makes proposed approach appealing as 
it can be applied to wide-variety of network setting and learning procedures. However, the equivalence would 
capture components of topological structure,  learning procedure and network setting. This sort of 
analysis can be considered as gaining understanding of structure and function relationships and 
stems from topological data analysis \cite{bubenik2015statistical, cang2017topologynet, rieck2018a, wasserman2018topological}.

The analysis of spectral density of weights of deep learning architectures recently 
investigated \cite{pennington18a, sagun18a, rieck2018a, martin19a, suezen19pse}. We follow 
a similar ethos in this regard, as earlier work is pioneered such analysis on investigating 
eigenvalue statistcs in neural network learning \cite{leCun91a}.

\section{Mixed matrix ensembles}

In establishing equivalence of two neural networks, we have taking a route that 
requires a mathematical setting in the language of {\it matrix ensembles}, square matrices. 
Matrix ensembles especially appear in random matrix theory \cite{mehta, wigner1967a}.
One of the prominent example is circular complex matrix ensembles \cite{dyson62a} that
mimics quantum statistical mechanics systems \cite{haake2013a}. It is hinted out that decay of spectral
ergodicity with increasing matrix order $N$ for circular matrix ensembles
signifies an analogous behaviour as using deeper layers in neural networks \cite{suzen17a},
whereby circular ensembles used as a simulation tool. However, real deep learning architectures have 
rarely all the same order weight matrices, but the weight matrices extracted from trained 
deep learning architectures will have variety of different orders due to different 
units in layer connections and forms a {\it Layer Matrix Ensemble}, 
see Definition \ref{def:lme}.

\begin{definition}
{\it Layer Matrix Ensemble} $\mathscr{L}^{m}$ \cite{suezen19pse}
The weights $W_{l} \in \mathbb{R}^{p_{1} \times p_{2} \times ... \times p_{n}}$ are obtained from
a trained deep neural network architecture's layer $l$ as an $n$-dimensional Tensor. A Layer
Matrix Ensemble $\mathscr{L}^{m}$ is formed by transforming $m$ set of weights $W_{l}$ to square matrices
$X_{l} \in \mathbb{R}^{N_{l} \times N_{l}}$, that $X_{l} = A_{l} \cdot A_{l}^{T}$ and
$A_{l} \in \mathbb{R}^{N_{l} \times M_{l}}$ is marely a stacked up version (projecting an arbitrary tensor to a matrix)  of
$W_{l}$ where $n > 1$, $N_{l}=p_{1}$, $M_{l}= \prod_{j=2}^{n} p_{j}$ and
$p_{j},n, m, N_{l}, M_{l}, j \in \mathbb{Z}_{+}$. Consequently $\mathscr{L}^{m}$ will
have $m$ potentially different $N_{l}$ size square matrices $X_{l}$ of at least
size $2 \times 2$.
\label{def:lme}
\end{definition}

{\it Circular Ensemble} is formed by drawing a complex circular
matrices from different size Circular Unitary Ensembles (CUEs), matching the orders,
$m$ different orders coming from $\mathscr{L}^{m}$ and taking their modulus, 
as we are dealing with real matrices in conventional deep learning architectures, see Definition \ref{def:CCUE}.

\begin{definition}
{\it Circular Unitary Mixed Ensemble}  $\mathscr{U}^{m}$ \\
Set of matrices $A_{l}=Mod(U_{l})$ where 
$U\in \mathbb{C}^{N_{l} \times N_{l}}$ and $A\in \mathbb{R}^{N_{l} \times N_{l}}$
forms this ensemble.  In component form, each $U_{l}$ obeys the following construction \cite{mezzadri06a, berry13a}:
Consider a Hermitian matrix $H \in \mathbb{C}^{N \times N}$,
$$H_{ij} = \frac{1}{2}(a_{ij}+ I b_{ij}+a_{ji}- I b_{ji}),$$
where $1 \leq i,j \leq N$, and $a_{ij}, b_{ij},
a_{ji}, b_{ji} \in \mathbb{G}$, i.e, they are elements of
the set of independent identical distributed Gaussian random
numbers sampled from a normal distribution and $I$ is
the imaginary number.

Ensemble matrices $U$ is defined as
$$U = exp(\gamma_{i} I) \cdot v_{j}^{i},$$
$v_{i}$ is the $i$-th eigenvector of $H_{ij}$,
where $\gamma_{i} \in [0, 2\pi]$ is a uniform random number.

\label{def:CCUE}
\end{definition}

Both $\mathscr{L}^{m}$ and  $\mathscr{U}^{m}$ forms a 
{\it Mixed Matrix Ensembles} (MMEs) as defined in Definition \ref{def:mme}.

\begin{definition}
{\it Mixed Matrix Ensembles} (MMEs) $\mathscr{M}^{m}$ are defined as set 
of $m$ square matrices $A_{i} \in \mathbb{R}^{N_{i} \times N_{i}}$, $i=1,..,m$,
where by $N_{i} \ge 2$ and $m,i,N_{i} \in \mathbb{Z}_{+}$. Mixed here implies set of
different size real square matrices forming an ensemble. In the case of all $N_{i}$ having the
same value makes MMEs a {\it pure} matrix ensemble. $N_{i}$ is sometimes called the 
order of a matrix as well.
\label{def:mme}
\end{definition}

\begin{figure}
  \includegraphics[width=0.4\textwidth]{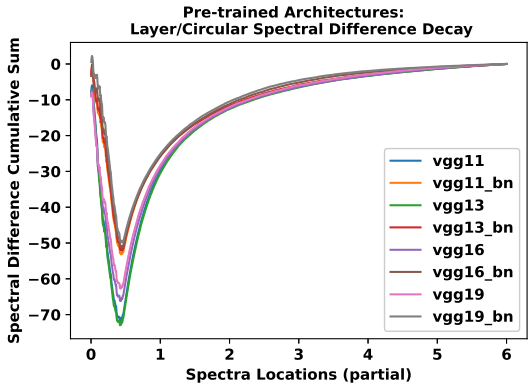}
  \caption{\label{fig:vgg} Circular Spectral Difference for vgg architectures. No smoothing, raw histograms. (color online)}
\end{figure}

\section{Conjugacy and Equivalence}

We introduce a concept of mixed matrix conjugate ensembles inspired from statistical physics \cite{costeniuc2005a, suzen2009a}.
Conjugacy of statistical mechanics ensembles are well founded based on Legendre Transforms \cite{costeniuc2005a, suzen2009a}.
Here we need to follow a different approach based on core characteristic of a matrix ensembles. There is no known
conjugacy rules for such ensembles and we introduce here one of the many possible conjugacy constructions. 

\begin{figure}
  \includegraphics[width=0.4\textwidth]{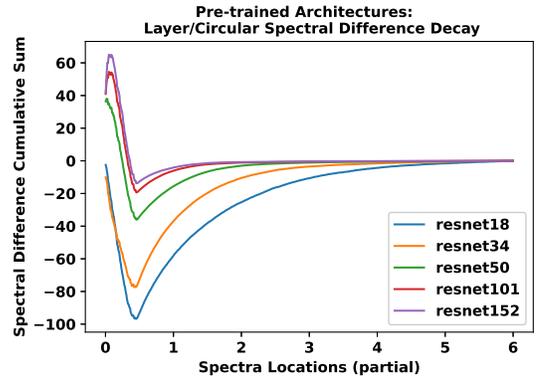}
  \caption{\label{fig:resnet} Circular Spectral Difference for resnet architectures. No smoothing, raw histograms. (color online)}
\end{figure}

Finding an appropriate conjugate mixed matrix ensemble given $\mathscr{L}^{m}$, layer matrix ensemble coming 
from a trained deep learning architecture or it could be synaptic network weight matrices from Brain networks 
for example \cite{rajan2006b, bullmore2009a} lies in pooled eigenvalues and spectra, see Definition \ref{def:pooleigen}.
This is a natural way of thinking conjugacy while core properties of matrices usually lies in spectral 
information of matrix group operators \cite{mudrov07q}.

\begin{definition}
{\it The pooled eigenvalues and spectra} of mixed matrix ensemble $\mathscr{M}^{m}$ is build from
the collection of eigenvalues of $m$ square matrices $A_{i} \in \mathbb{R}^{N_{i} \times N_{i}}$, $i=1,..,m$,
denoted by $\epsilon_{j}^{pool}$ and $j=1,...,N_{i} \cdot m$, with a corresponding spectral density
$\rho^{pool}(\epsilon)$.
\label{def:pooleigen}
\end{definition}

Now, we have well defined setting for defining conjugate ensemble and equivalence condition for MMEs.
These are summarized in Definitions \ref{def:conj} and \ref{def:eq}.

\begin{definition} {Conjugate MMEs} \\
Given two mixed matrix ensembles, $\mathscr{M}_{1}^{m}$ and $\mathscr{M}_{2}^{m}$, forms
a conjugate ensembles if their respective cumulative spectral density difference approaches to zero 
over long positive tail part of the spectrum, Hence at $[0.0, \epsilon_{max}]$, Circular Spectral Difference (CSD),
$\Delta_{CSD}(\epsilon) = \rho_{1}(\epsilon) - \rho_{2}(\epsilon)$ and with cumulative CSD defined as
$\sum_{0}^{\epsilon_max} \Delta_{CSD}(\epsilon)$ over spectral locations approaches to zero for large enough $\epsilon_{max}$, 
The $\epsilon_{max}$ being at least the largest eigenvalue to consider in constructing the spectral density.
\label{def:conj}
\end{definition}

\begin{definition} {Equivalence of MMEs} \\
Given two mixed matrix ensembles $\mathscr{M}_{1}^{m}$ and $\mathscr{M}_{2}^{m}$ are equivalent if
following two conditions met: \\
1. There is a third mixed matrix ensembles $\mathscr{M}_{c}^{m}$ that is conjugate to both. \\
2. The variance of  Circular Spectral Difference (CSD) of them are equivalent with a small $\delta$, \\
$Var \left( \Delta_{CSD}^{1} \right) -  Var \left( \Delta_{CSD}^{2} \right) = \delta$ \\
\label{def:eq}. The choice of $\delta$ is an engineering decision that architecture search should decide.
\end{definition}

\begin{figure}
  \includegraphics[width=0.4\textwidth]{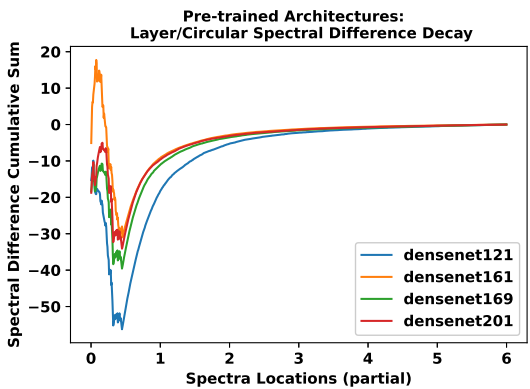}
  \caption{\label{fig:densenet} Circular Spectral Difference for densenet architectures. No smoothing, raw histograms. (color online)}
\end{figure}

\section{Experiments with vision architectures}

\begin{figure}
  \includegraphics[width=0.4\textwidth]{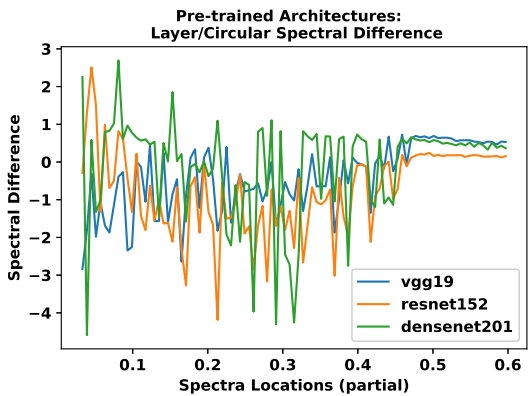}
  \caption{\label{fig:fluc} Circular Spectral Difference fluctuations for vgg99, resnet152 and densenet201. (color online)}
\end{figure}

Our experiments focus on deep neural network architectures developed for vision task. Three different type 
of vision architectures are used: VGG \cite{simonyan14a}, ResNet \cite{he2016a}, and DenseNet \cite{huang17a} 
with different depth and batch normalisation for VGG. We generated mixed Layer Matrix Ensembles for all mentioned 
type networks using pretrained weigths \cite{pytorch} and their corresponding mixed Unitary Circular Ensembles as a proposal
conjugate ensembles. We used positive range for long positive tail part of the spectrum $[0,0, 6.0]$ with $1000$ equalspacing. 
This provides sufficiently smooth data with the given pooled eigenvalues from produced ensembles.

Computed CSDs for long positive tail part is shown for VGG, ResNet and DenseNet architectures on Figures \ref{fig:vgg}, \ref{fig:resnet}
and \ref{fig:densenet} respectively. We observe that different architecture's CSDs decay in different rates. This indicates different 
fluctuation characteristics of each CSDs for different architectures as demonstrated on Figure \ref{fig:fluc}.

Variance of all CSDs for all investigated architectures are summarized in Table \ref{table:var}. We clearly see that 
each architecture type clearly seperated by Variance of CSDs. We see the equivalance of all VGG architectures implying
the reported performance metrics are not too far apart in reality, VGG with batch normalisation architectures forms the subset, this is also
observed in cumulative CSDs in Figure \ref{fig:vgg}.  However the equivalance goes away very quickly with increasing depth, this is 
observed between resnet18 and resnet152, similarly between densenet121 and densenet201. Equivalence between densenet169 and densenet201 
could also be accepted.

\section{Conclusions}

We have developed a methematically well defined approach to detect equivalance between deep learning architectures. The method
relies on building conjugate matrix ensembles and investigating their spectral difference over the long positive tail part.
This approach can also be used in detecting Brain motifs if synaptic weights are known. 

The method is very practical and can be used in designing new artifical neural architectures via Neural Architecture Search (NAS) or 
compression of the existing known architecture by systematic or random reduction of the network size and computing variance of CSD as proposed 
in this work. The approach here can be thought as a complexity measure for architecture groups rather than the fine grain 
complexity measure for a single network, that the group of networks would have similar test performances, hence will be useful
in establishing equivalance classes for deep neural networks.. 

\begin{table}[]
\centering
\begin{tabular}{|r|r|r|r|}
\hline
Architecture  & Top-1 error  &  Top-5 error  & Variance CSD   \\ \hline
vgg11         & 30.98        &   11.37       &  0.19   \\ \hline
vgg13         & 30.07        &   10.75       &  0.20   \\ \hline
vgg16         & 28.41        &    9.63       &  0.19   \\ \hline
vgg19         & 27.62        &    9.12       &  0.18   \\ \hline
vgg11bn       & 29.62        &   10.19       &  0.10   \\ \hline
vgg13bn       & 28.45        &    9.63       &  0.09   \\ \hline
vgg16bn       & 26.63        &    8.50       &  0.10   \\ \hline
vgg19bn       & 25.76        &    8.15       &  0.09   \\ \hline
resnet18      & 30.24        &   10.92       &  0.20   \\ \hline
resnet34      & 26.70        &    8.58       &  0.23   \\ \hline
resnet50      & 23.85        &    7.13       &  1.45   \\ \hline
resnet101     & 22.63        &    6.44       &  1.86   \\ \hline
resnet152     & 21.69        &    5.94       &  1.98   \\ \hline
densenet121   & 25.35        &    7.83       &  0.42   \\ \hline
densenet161   & 22.35        &    6.20       &  0.29   \\ \hline
densenet169   & 24.00        &    7.00       &  0.52   \\ \hline
densenet201   & 22.80        &    6.43       &  0.54   \\ \hline
\end{tabular}
\
\caption{Variance of CSD per architecture corresponding Top-1 and Top-5 classification errors on ImageNet dataset.}
\label{table:var}
\end{table}

\begin{acknowledgments}
Authour is grateful for PyTorch \cite{pytorch} team's superb work on bundling pretrained architectures as easily accessible
modules and providing Top-1 and Top-5 errors in a concise manner. I express my gratitute to Nino Malekovic (originally of ETH Zurich) 
for his feedback, pointing out to me several things: connections to the  topological data analysis literature, 
suggesting justification of conjugacy and bounds of delta, 
in CSD variances.
\end{acknowledgments}

\section*{Supplementary material}
A Python code notebook with functions to reproduce the data and results is provided with this manuscript, {\it deep\_dyson\_networks.ipynb}.

\bibliography{dyson}

\end{document}